\documentclass[10pt,twocolumn,letterpaper]{article}

\usepackage{cvpr}              

\usepackage{multirow}

\definecolor{cvprblue}{rgb}{0.21,0.49,0.74}
\usepackage[pagebackref,breaklinks,colorlinks,allcolors=cvprblue]{hyperref}
\usepackage{booktabs}
\usepackage{makecell}

\def\paperID{} 
\def\confName{CVPR}
\def\confYear{2026}

\title{PhenoYieldNet: Learning Crop-Aware Phenological Responses for\\ Multi-Crop Yield Prediction}

\author{
Yu Luo$^{1}$ \quad
Xiaogang Zhu$^{2, *}$ \quad
Shan Zeng$^{3}$ \quad
Wei Xiang$^{4}$ \quad \\
Thomas Francis Bishop$^{1}$  \quad 
Zhiyong Wang$^{1}$\quad
Kun Hu$^{5,}$\thanks{Corresponding authors.} \\[4pt]
$^{1}$School of Computer Science, The University of Sydney, NSW 2006, Australia \quad \\
$^{2}$ School of Computer Science and Information Technology, Adelaide University, SA 5005, Australia \\
$^{3}$ College of Mathematics and Computer Science, Wuhan Polytechnic University, Wuhan, China \\
$^{4}$ School of Computing, La Trobe University, NSW 2000, Australia\\
$^{5}$ School of Science, Edith Cowan University, WA 6027, Australia\\
{\tt\small \{yluo0465, thomas.bishop, zhiyong.wang\}@sydney.edu.au,} \quad
{\tt\small xiaogang.zhu@adelaide.edu.au} \quad \\
{\tt\small zengshan1981@whpu.edu.cn} \quad
{\tt\small w.xiang@latrobe.edu.au} \quad
{\tt\small k.hu@ecu.edu.au} \\
}

\begin{document}
\maketitle
\begin{abstract}
Accurate crop yield prediction is crucial for sustainable agriculture and global food security.
While existing methods are predominantly developed for single-crop prediction, they often struggle to generalize across diverse crop types, without addressing the unique crop phenological responses that are dynamically modulated by complex weather patterns. 
In this paper, we propose PhenoYieldNet, a multi-crop yield prediction framework that learns crop-specific phenology by explicitly modeling their responses with temporal drivers. Specifically, we develop a crop-aware temporal decoder consisting of a Crop Phenology Bank (CPB) and a Crop Phenology Attention (CPA) module. The CPB integrates a set of learnable embeddings, which leverage a query to guide the CPA module to learn the most relevant phenology patterns for the specific crop. And the CPA module explicitly captures multi-scale trend and variation components to construct temporal contexts, enabling the model to dynamically adjust the attention across different phenological stages.
To learn robust and generalizable features for multi-crop prediction, the encoder is initialized with a pre-trained foundation model, and further adapted via a self-supervised Temporal Contrastive Adaptation strategy to align with agricultural temporal dynamics.
Extensive experiments conducted on multi-crop datasets indicate that our proposed method significantly outperforms state-of-the-art methods, exhibiting strong generalization capabilities across different regions and crops.
Code is available at \url{https://github.com/roroyo/PhenoYieldNet}
\end{abstract}
\section{Introduction}
\label{sec:intro}

\begin{figure}[t]
\centering
\includegraphics[width=1.0\linewidth]{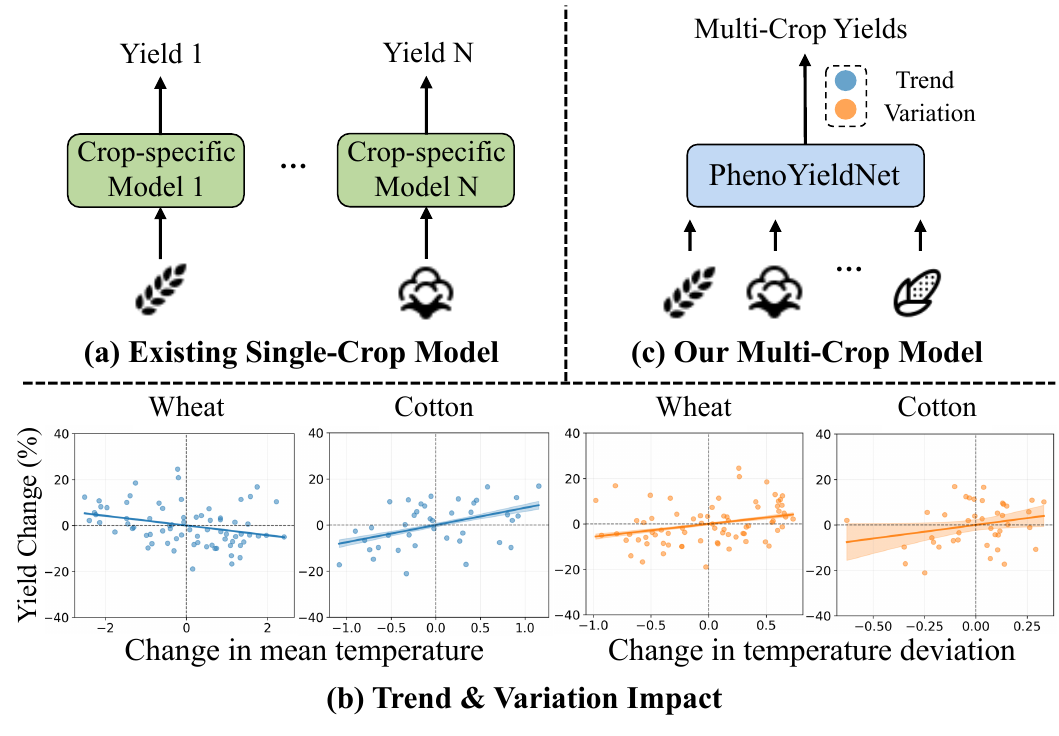}
\caption{(a) Existing yield prediction methods typically train a separate model for each individual crop.
(b) Empirical evidence shows that different crops exhibit complex and distinct responses to meteorological variables.
(c) Our proposed PhenoYieldNet, a multi-crop yield prediction framework, explicitly models crop-specific phenological patterns with meteorological variations.}
\label{Fig1_contri}
\vspace{-20pt}
\end{figure}

Global climate change presents unprecedented challenges to agricultural systems and food security~\cite{rezaei2023climate}. With shifting patterns and frequent extreme weather events, accurate crop yield prediction has attracted increasing attention in agriculture~\cite{he2023physics}. It is very helpful for farmers, breeding organizations, and government agencies to make informed decisions, ultimately optimizing yield outcomes and ensuring long-term agricultural stability and sustainability~\cite{qin2024global}.

Recent advances in remote sensing have provided abundant earth observation data throughout entire crop growing cycles, from satellite imagery that tracks crop development status to meteorological data that records the dynamic environmental factors influencing crop growth. 
These methods~\cite{lu2024goa,sun2019county,lin2023mmst,mena2025adaptive} are predominantly developed for individual crops, trained within crop-specific and region-specific settings (Fig.~\ref{Fig1_contri}(a)). While effective in isolated cases, such methods struggle to generalize across different crops or adapt to unseen regions. However, crops grown in different parts of the world often face similar conditions—such as environmental factors, nutrient deficiency and relevant trends—and may exhibit both shared and distinct yield responses.
Modeling these cross-crop commonalities and differences in a unified framework can enable knowledge transfer, improve prediction for data-scarce crops, and improve robustness to regional variability.
This raises a fundamental question:
\textit{Can we build a unified model that generalizes across diverse crops, while capturing both shared patterns and crop-specific characteristics?}

Developing a unified model for multi-crop yield prediction poses two key challenges: (1) capturing crop-aware phenological characteristics, and (2) modeling how weather variability influences phenology and, in turn, affects yield. Specifically, recent findings in agriculture~\cite{hu2025impact, rezaei2023climate, nyborg2022generalized} highlight both long-term trends (e.g., gradual warming) and short-term extremes (e.g., droughts, heatwaves) as key drivers of crop phenology and yield. 
Yet, existing yield prediction methods often merge satellite and meteorological modalities in a straightforward manner, without explicitly modeling how meteorological conditions affect phenology at different growth stages~\cite{chang2024target, gopi2024red}.
Empirical analysis further reveals that different crops exhibit distinct responses to both long-term trends and short-term variability. For instance, considering the temperature variable, wheat and cotton yields show opposite responses to mean temperature (trend), and differ in their sensitivity to the standard deviation of temperature (variation), as illustrated in Fig.~\ref{Fig1_contri}(b).

Therefore, we propose PhenoYieldNet, a novel framework for multi-crop yield prediction. It learns distinct crop-specific phenological responses to temporal trends and variations.
At the core of PhenoYieldNet is a crop-aware temporal decoder that integrates two key components: the Crop Phenology Bank~(CPB) and the Crop Phenology Attention~(CPA).
The CPB formulates crop-specific phenology patterns into a set of learnable embeddings, which are used to guide the CPA in attending to the most relevant temporal features for each crop.
To capture how temporal trends and variations influence yields, the CPA module first explicitly decomposes the learned feature sequences into multi-scale trend and variation components. The information is then used as a context to refine temporal attention, allowing the model to dynamically weight the crop-specific phenological sensitivities.

PhenoYieldNet is trained in two stages to incorporate both general remote sensing knowledge and crop dynamics.
In the first stage, the encoder is initialized from a remote sensing foundation model and further adapted using a self-supervised Temporal Contrastive Adaptation (TCA) strategy, enhancing the model’s sensitivity to agricultural temporal patterns.
In the second stage, the encoder is frozen, and only the decoder is fine-tuned in a supervised manner for yield prediction. Extensive experiments demonstrate that PhenoYieldNet achieves state-of-the-art performance compared to existing single-crop and multi-crop yield prediction methods.

Our major contributions are summarized as follows:

\begin{itemize}
\item We propose PhenoYieldNet, a unified framework for multi-crop yield prediction that captures both shared agricultural patterns and crop-specific phenological responses within a single model.

\item We design a crop-aware temporal decoder that integrates a Crop Phenology Bank and a Crop Phenology Attention module, enabling the model to adaptively attend to critical temporal variations based on crop-specific phenology.

\item We introduce a Temporal Contrastive Adaptation strategy to bridge the gap between general-purpose remote sensing knowledge and agricultural temporal dynamics, improving model generalization. 

\item Experiments on real-world datasets show that PhenoYieldNet outperforms state-of-the-art single-crop and multi-crop yield prediction methods.
\end{itemize}

\section{Related Work}
\label{sec:related work}

\paragraph{Crop Yield Prediction.}
Early crop yield prediction methods often relied on a single data source.
One line of research focuses on meteorology-based prediction, which models the direct influence of weather variables on crop yield \cite{khaki2020cnn, fan2022gnn, jeong2016random, wen2024duocast,wen2026mccast,wen2026stable}.
Another line leverages remote sensing~(RS) imagery, which captures phenological changes in crops (e.g., temporal variations in vegetation indices) \cite{ju2021optimal, yadav2025context, dai2025vg}.
To extract spatio-temporal features from RS data, various architectures have been employed, including convolution neural networks~(CNNs)~\cite{nevavuori2019crop, baghdasaryan2022deep}, recurrent neural networks~(RNNs)~\cite{chang2024target}, and hybrid CNN–RNN models~\cite{mao2020context}.
For instance, UNet-ConvLSTM was proposed to capture multi-scale spatial features and temporal dependencies in satellite image sequences~\cite{kamangir2024large}.

Yet, crop yield is a complex outcome driven by both crop growth dynamics, as captured by RS data, and meteorological factors.
Recent studies increasingly integrate multiple modalities via joint encoding and fusion~\cite{you2017deep,sun2019county,kaur2022generalized, dai2024one}.
For example, 
MMST-ViT~\cite{lin2023mmst} extracted features from satellite imagery and meteorological data using a Transformer and an MLP, then fuses them with a cross-modal Transformer.
Recognizing that different modalities contribute unequally, MMVF~\cite{mena2025adaptive} was proposed to adaptively fuse multi-source data with gating mechanisms.

Despite these advancements, most existing methods remain single-crop focused.
To the best of our knowledge, the only work addressing multi-crop yield prediction is YieldNet~\cite{khaki2021simultaneous}, which employs a shared CNN backbone with two separate prediction heads for corn and soybean.
While this represents a step towards multi-crop modeling, it does not fully explore the complex and dynamic interactions between meteorological drivers and crop-specific phenological responses, which are essential for generalizable multi-crop yield prediction.

\paragraph{Spatio-Temporal Understanding in RS.}
RS technologies provide rich multi-temporal Earth Observations, enabling detailed monitoring and understanding in downstream tasks.
Early deep learning approaches sought to extract spatio-temporal features by jointly modeling spatio-temporal patterns.
Hybrid architectures, such as CNN–RNNs~\cite{mao2020context}, process spatial features sequentially, while unified models like 3D CNNs~\cite{nejad2022multispectral} and Temporal Convolutional Networks~\cite{mohan2023temporal} learn spatio-temporal representations holistically.
Subsequently, Transformer-based models have shown strong potential in capturing long-range temporal dependencies and in identifying key phenological stages~\cite{liu2022rice, gallo2024enhancing, guo2024novel, shi2025culture,dou2026beyond}.

More recently, the emergence of Remote Sensing Foundation Models (RSFMs)~\cite{wang2022empirical, guo2024skysense, cong2022satmae, noman2024rethinking, hong2024spectralgpt, su2025ri} has led to a paradigm shift in RS data processing.
By pre-training on large-scale unlabeled imagery, RSFMs learn general-purpose representations that can be transferred across downstream tasks.
However, their application to crop yield prediction remains limited, as RSFMs are not trained to capture visual representations of RS data together with crop-specific phenology information and meteorological dynamics.
Our goal is to bridge the gap between general-purpose RS representations and the specialized knowledge required for accurate multi-crop yield prediction by incorporating domain-specific agricultural insights.

\section{Proposed Method}

\begin{figure*}[tp]
\centering
\includegraphics[width=1.0\linewidth]{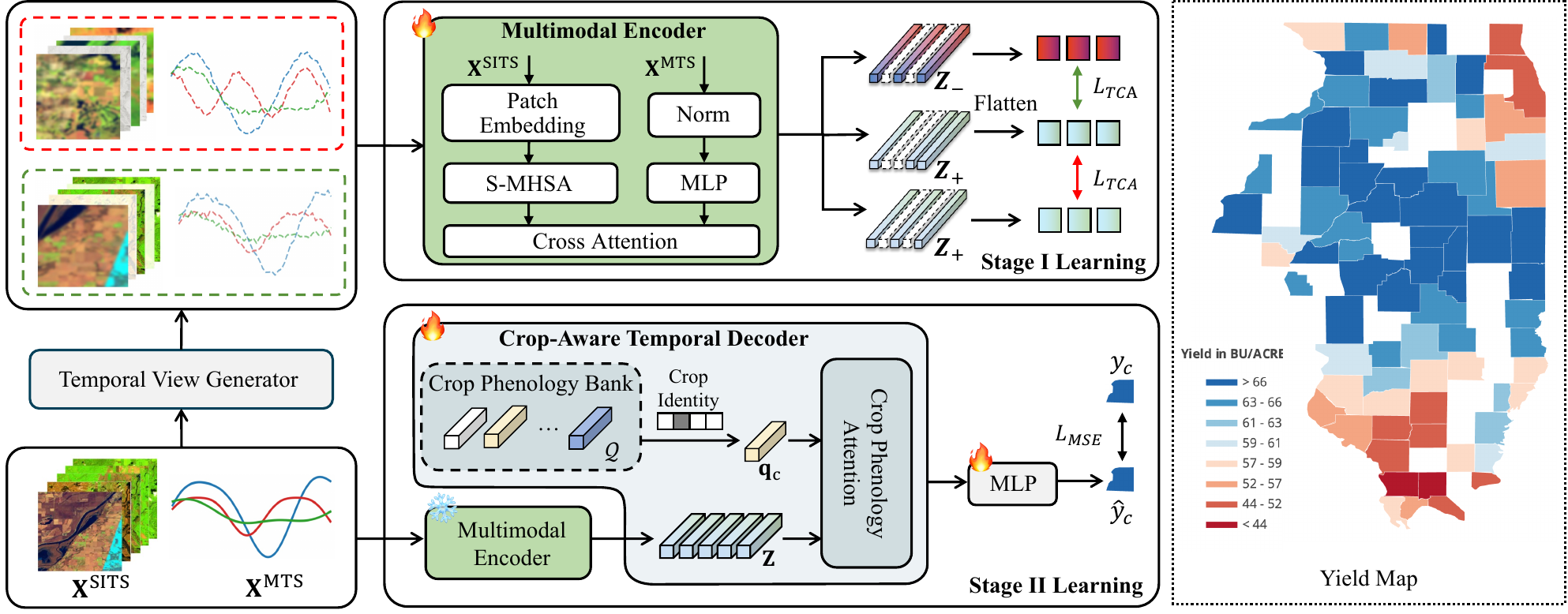}
\caption{
Overview of the PhenoYieldNet architecture.
A multimodal encoder processes satellite image time series ($\mathbf{X}^{\text{SITS}}$) and meteorological time series ($\mathbf{X}^{\text{MTS}}$) through separate branches and fuses them via a cross-attention module.
The resulting representation is passed to a crop-aware temporal decoder, which integrates a Crop Phenology Attention module guided by a Crop Phenology Bank to model crop-specific temporal responses.
An MLP head is used for yield regression.
PhenoYieldNet is trained in two stages: (I) temporal contrastive pre-training of the encoder with a remote sensing foundation model, and (II) supervised fine-tuning of the decoder for yield prediction.
}
\label{Fig2}
\end{figure*}

\subsection{Overview \& Problem Formulation}
\label{overview}

We consider an input consisting of two components: (1) satellite image time series~(SITS) $\mathbf{X}^{\text{SITS}}\in \mathbb{R}^{T \times H \times W \times B}$, where $T$ is the number of time steps, and $H$, $W$, and $B$ denote the image height, width, and number of spectral bands, respectively; 
and (2) meteorological time series $\mathbf{X}^\text{MTS} \in \mathbb{R}^{T\times N_d \times M}$, where $N_d$ indicates the number of higher-frequency records within each satellite observation period (e.g., $T$ for weeks and $N_d$ for days within a week), and $M$ is the number of weather variables. 
Multi-crop yield prediction aims to estimate the ground truth county-level yield $\mathbf{y}_c$ as $\hat{\mathbf{y}}_c$ for a given crop type $\mathbf{c} \in \{ 1, \cdots,C \}$.

As illustrated in Fig.~\ref{Fig2}, PhenoYieldNet adopts an encoder-decoder framework, which consists of a multimodal encoder and a crop-aware temporal decoder. The multimodal encoder first extracts latent features from $\mathbf{X}^{\text{SITS}}$ and $\mathbf{X}^\text{MTS}$ via a ViT and an MLP, respectively, and then fuses them to produce a unified feature sequence $\mathbf{Z} \in \mathbb{R}^{T \times d}$, where $d$ denotes the hidden dimension. Next, the crop-aware temporal decoder takes the latent features as input to model crop-specific phenological dynamics, enabling accurate yield prediction.
Let $\mathcal{E}$ and $\mathcal{D}$ denote the encoder and decoder, respectively. The overall model architecture can be formulated as:
\begin{equation}
    \mathbf{Z} = \mathcal{E}(\mathbf{X}^{\text{SITS}}, \mathbf{X}^\text{MTS}),
\end{equation}
\begin{equation}
    \hat{\mathbf{y}}_c = \mathcal{D}(\mathbf{Z}).
\end{equation}

The entire framework is trained in two stages. In the first stage, we employ a temporal contrastive adaptation strategy to continually pre-train the multimodal encoder to adapt to the specific domain of crop yield prediction. The details of this strategy are introduced in the following section. In the second stage, the pretrained encoder is frozen, and only the crop-aware temporal decoder and the prediction head are fine-tuned in a supervised manner using the mean squared error (MSE) loss:
\begin{equation}
    L_{\text{MSE}} = \frac{1}{N}\sum_{i=1}^{N} \| \log\hat{\mathbf{y}}_c^{(i)} - \log \mathbf{y}_c^{(i)} \|_2^2,
\end{equation}
where $\mathbf{y}_c^{(i)}$ is the ground-truth yield for the $i$-th sample, and $N$ is the total number of samples. To mitigate the impact of scale variability across different crop types, the loss is computed in the logarithmic space and transformed back to the original scale using the exponential function.

\subsection{Multimodal Encoder}
\label{ME}

The multimodal encoder aims to extract and fuse spatial and temporal features from heterogeneous modalities.
For the satellite image time series, we adopt a vision Transformer to encode visual features. Inspired by the strong transfer learning capability of foundation models, we initialize the encoder using weights from a pre-trained RSFM, SpectralGPT~\cite{hong2024spectralgpt}.
For the meteorological time series, we employ an MLP as an adapter to project the raw data at each time step into a corresponding embedding.
To effectively capture complex relationships between satellite imagery and meteorological signals, a cross-attention module is adopted to generate fused multimodal representations.

\subsection{Crop-Aware Temporal Decoder}
\label{CATD}

To learn the distinct phenological patterns of various crops from a common encoded latent space, we propose a unified Crop-Aware Temporal Decoder, built upon a Temporal Transformer. It comprises two key components: a \textit{Crop Phenology Bank} and a \textit{Crop Phenology Attention} module, both of which are detailed in the following subsections.

\subsubsection{Crop Phenology Bank.}
\label{CPB} 
To enable the decoder to capture crop-specific phenological patterns from the shared multimodal latent space, we introduce a crop phenology bank (CPB) - a set of learnable vectors, each representing the typical phenological characteristics of a specific crop type. 
Formally, the crop phenology bank is defined as:

\begin{equation}
    \mathcal{Q} = \{ \mathbf{q}_c \in \mathbb{R}^{d} \mid c\in C \}, 
\end{equation}
where $\mathbf{q}_c$ is the query vector for crop $c$, and $d$ is the hidden dimension. The query vectors are randomly initialized from a normal distribution $N(0,1)$ independently.
 
As illustrated in Fig.~\ref{Fig2}, during decoding, the decoder retrieves the crop-specific vector $\mathbf{q}_c$ as a query according to the crop identity. This query serves as a proxy for the crop's phenological signature and is utilized in crop phenology attention blocks to guide the interaction with the encoded multimodal feature embedding $\mathbf{Z}$. In this way, the decoder is encouraged to focus on temporal features that are most relevant to the phenological development of relevant crop. 


\subsubsection{Crop Phenology Attention.}
\label{CPA}

\begin{figure}[tp]
\centering
\includegraphics[width=0.87\linewidth]{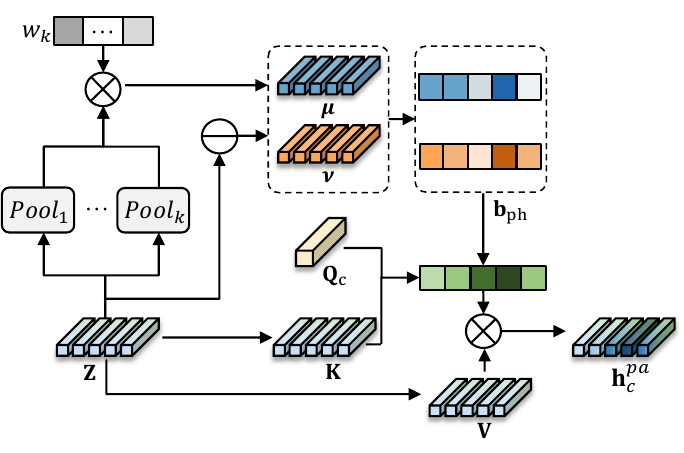}
\caption{Illustration of Crop Phenology Attention module. 
}
\label{fig3_decompose}
\end{figure}

To explicitly model temporal trends and phenological variations that affect crop yield, we introduce a crop phenology attention (CPA) module that leverages crop phenology queries to focus on the most informative stages via the phenology bank (shown in Fig.~\ref{fig3_decompose}). 

We first model the trends and variations across crop growth stages using a time-series decomposition strategy, inspired by \cite{zhou2022fedformer}.
Given that crop growth manifests across multiple temporal scales, ranging from short-term seasonal shifts to long-term annual trajectories, we employ multi-scale average pooling via sliding windows of varying lengths to capture the multi-scale trend patterns aligning with crop development cycles. 
These multi-scale patterns are then adaptively aggregated through learnable weights to yield the final trend component $\mu$, and the residual deviation is the variation component $\nu$.
Given the latent feature embedding $\mathbf{Z}$ covering a full year cycle as input, the process can be formulated as:
\begin{equation}
    \mu = \sum_{k \in \{3,6,12\}} w_k(\mathbf{Z}) \cdot\text{Pool}_{k}(\mathbf{Z}),
\end{equation}
    
\begin{equation}
    \nu = \mathbf{Z} - \mu,
\end{equation}
where $\text{Pool}_k(\cdot)$ denotes pooling with window size $k$, and $w_k(\cdot)$ is a learnable weight for aggregation. Noted that, padding operations are within $\text{Pool}_k(\cdot)$ to ensure the consistent dimension for outputs across temporal scales.

Next, both trend and variation components are projected as a bias vector $\mathbf{b}_{ph}$:


\begin{equation}
\begin{aligned}
    \textbf{b}_{ph} = \frac{1}{\sqrt{d}} \Big( & \lambda_{\mu}(\mathbf{W}^Q_{\mu} \mu) \cdot (\mathbf{W}^K_{\mu} \mu)^T \\
    & + \lambda_{\nu}(\mathbf{W}^Q_{\nu} \nu) \cdot (\mathbf{W}^K_{\nu} \nu)^T \Big),
\end{aligned}
\end{equation}
where $\lambda{\mu}, \lambda_{\nu}$ denotes the weight for the corresponding query and key matrices $\mathbf{W}_{\mu}$, $\mathbf{W}_{\nu}$ of trend and variation, respectively. 
Here, we specifically extract the global representation from the trend/variation attention matrices using a \texttt{[CLS]} token for $\mathbf{b}_{ph}$. 


Finally, the bias is injected into the phenology-guided attention, which can be formulated as:

\begin{equation}
    [\mathbf{Q}_c,\mathbf{K}, \mathbf{V}] = [\mathbf{W}^Q \mathbf{q}_c, \mathbf{W}^K \mathbf{Z},\mathbf{W}^V \mathbf{Z}],
\end{equation}

\begin{equation} 
    \mathbf{h}_c^{pa} = \sigma(\frac{\mathbf{Q}_c \mathbf{K}^T}{\sqrt{d}} + \textbf{b}_{ph} )\mathbf{V}, 
\label{phenologyattention}
\end{equation}
where $\mathbf{W}^Q$, $\mathbf{W}^K$ and $\mathbf{W}^V$ are learnable weights for linear projections, $\mathbf{h}_c^{pa}$ is the crop-aware feature embedding.

While the crop-specific query $\mathbf{q}_c$ allows the model to learn generally important phenological stages for the target crop, the bias $\mathbf{b}_{ph}$ further adjusts the focus on the critical stages influenced by unique trends and variations.


\subsection{Temporal Contrastive Adaptation}
\label{TCA}

The encoder in our PhenoYieldNet model is initialized from a foundation model pre-trained on general-purpose remote sensing data, enabling knowledge transfer from large-scale models. However, it remains insensitive to crop-aware phenological patterns and temporal dependencies that are critical for yield prediction. To bridge this gap, we introduce a temporal contrastive adaptation strategy - Learning Stage I - which aligns the pre-trained representations with the temporal dynamics of crop growth, thereby reducing the domain gap before proceeding to conventional supervised training - Learning Stage II.

We employ a temporal contrastive learning approach on multimodal time series inputs $\mathbf{X}^{\text{SITS}}$ and $\mathbf{X}^{\text{MTS}}$. 
Following standard contrastive learning, for each sample in a batch, we generate two different masked views. Each view is created by applying a randomly and independently generated mask $m \in \{ 0, 1\}^T$, where $T$ is the length of the image sequence. The masks are shared across modalities. Two views of the same sample (same location/year) form a positive pair, while other samples in the batch (from different locations) serve as negative pairs. 
For the positive pairs, we have corresponding encoded representations $\mathbf{Z}$ and $\mathbf{Z}_{+}$. 
A temporal contrastive loss is then applied as: 


\begin{equation}
L_{TCA} = -\log \frac{\exp(\text{sim}(\mathbf{Z}, \mathbf{Z}_{+})/\tau)}{\sum_{\mathbf{Z}^{'} \in {\mathbf{Z}+} \cup \{\mathbf{Z}_{-}\}}\exp(\text{sim}(\mathbf{Z}, \mathbf{Z}^{'})/\tau)},
\end{equation}
where $\{\mathbf{Z}^-\}$ is the set of all negative counterparts, $\tau$ is the temperature parameter, and $\text{sim}(\cdot, \cdot)$ is the cosine similarity.

\section{Experiments}

\subsection{Experimental Settings}
\subsubsection{Datasets}
In our experiments, we utilize two datasets from CropNet~\cite{lin2023mmst} and \cite{you2017deep} (denoted as MODIS here). The CropNet dataset contains Sentinel-2 satellite imagery and meteorological data from HRRR, covering the years from 2017 to 2022. It includes four major crop types—soybean, corn, cotton, and winter wheat—that together represent a significant portion of U.S. agricultural production. 
The MODIS dataset includes both imagery and meteorological data from the MODIS satellite, spanning the years from 2003 to 2015, including a single crop type: corn. Both datasets cover 11 U.S. states, and the ground-truth, yearly average  crop yield labels at county-level are sourced from the USDA statistics database. Leveraging these datasets allows us to evaluate our model’s generalization capability across diverse crop types and locations. 
Following \cite{lin2023mmst}, we train the model on the data of CropNet from 2020 and MODIS from 2014, while testing on the data from the next year. As CropNet splits county-level imagery into $G$ sub-regions, we follow \cite{lin2023mmst} that uses a spatial Transformer to aggregate the sub-regions with the \texttt{[cls]} tokens.

To focus the analysis on critical growing stages, only the periods corresponding to each crop's growing season are preserved. More details are provided in Supplementary Material.

\subsubsection{Comparison Methods}

We compare our proposed method with two categories of baseline approaches:
(a) Single-crop yield prediction methods, including machine learning-based methods: random forest~\cite{breiman2001random}, XGBoost~\cite{chen2016xgboost}; and deep learning-based methods: MMST-ViT~\cite{lin2023mmst}, a Transformer-based method~\cite{helber2024operational}, UNet-ConvLSTM~\cite{kamangir2024large}, and MMVF~\cite{mena2025adaptive};
(b) Multi-crop yield prediction method: YieldNet~\cite{khaki2021simultaneous}.
Detailed descriptions of these methods are provided in Supplementary Material.

\begin{table}[tp]
\centering
\begin{tabular}{lccc}
\toprule
Method        & RMSE  $\downarrow$ & $R^2$  $\uparrow$ & Corr  $\uparrow$ \\ \midrule
MMST-ViT~\citeyearpar{lin2023mmst}     & 8.12    & 0.400                   & 0.632   \\
Transformer~\citeyearpar{helber2024operational}  & 9.31    & 0.170                   & 0.412   \\
Unet-ConvLSTM~\citeyearpar{kamangir2024large}& \underline{6.33}    & \underline{0.586}                   & \underline{0.766}   \\
MMVF~\citeyearpar{mena2025adaptive}        & 10.13   & 0.260                   & 0.510    \\
PhenoYieldNet (Ours)      & \textbf{5.95}    & \textbf{0.663}                   & \textbf{0.814}   \\ \bottomrule
\end{tabular}
\caption{Performance comparison of yield prediction on the MODIS dataset. 
The best and second-best results are highlighted in \textbf{bold} and with an \underline{underline}, respectively.
}
\label{table_modis_baseline}
\end{table}

\begin{table*}[htbp]
\setlength{\tabcolsep}{0.7mm}
\centering
\begin{tabular}{lcccccccccccc}
\toprule
\multirow{2}{*}{Method} & \multicolumn{3}{c}{Corn}                                  & \multicolumn{3}{c}{Cotton}                                & \multicolumn{3}{c}{Soybean}                              & \multicolumn{3}{c}{Winter Wheat}                         \\ \cmidrule(lr){2-4} \cmidrule(lr){5-7} \cmidrule(lr){8-10} \cmidrule(lr){11-13}
                        & RMSE $\downarrow$        & $R^2$ $\uparrow$ & Corr $\uparrow$        & RMSE $\downarrow$      & $R^2$ $\uparrow$ & Corr $\uparrow$        & RMSE $\downarrow$     & $R^2$ $\uparrow$ & Corr $\uparrow$        & RMSE $\downarrow$     & $R^2$ $\uparrow$ & Corr $\uparrow$        \\ \midrule
RF~\citeyearpar{breiman2001random}                 & 21.96          & 0.408                   & 0.639          & 85.32          & 0.174                   & 0.417          & 6.76         & 0.537            &0.732	       & 10.98         & 0.036               & 0.190           \\
XGBoost~\citeyearpar{chen2016xgboost}                & 22.72        & 0.369              & 0.607           & 84.47       & 0.166                     & 0.407         & 7.59       & 0.432          &0.657  	       & 10.68        & 0.038              & 0.195          \\
MMST-ViT~\citeyearpar{lin2023mmst}                & 25.98          & 0.215                   & 0.464          & 88.25          & 0.468                   & 0.684          & 6.58          & 0.590                   & 0.768          & 10.16         & 0.146                   & 0.382          \\
Transformer~\citeyearpar{helber2024operational}             & 34.35          & 0.225                   & 0.474          & 405.53         & 0.239                   & 0.488          & 8.89          & 0.352                   & 0.593          & 23.51         & 0.145                   & 0.381          \\
Unet-ConvLSTM~\citeyearpar{kamangir2024large}           & 31.34          & 0.381                   & 0.617          & 241.97         & \underline{0.540}          & \underline{0.735} & 9.74          & 0.442                   & 0.665          & 14.42         & \underline{0.167}                   & \underline{0.409}          \\
MMVF~\citeyearpar{mena2025adaptive}                    & 71.66          & 0.181                   & 0.426          & 457.93         & 0.293                   & 0.541          & 15.11         & 0.200                   & 0.447          & 32.50         & 0.075                   & 0.273          \\
PhenoYieldNet-SC (Ours)            & \underline{17.27} & \underline{0.493}          & \underline{0.667} & \underline{71.22} & 0.526                   & 0.725          & \underline{6.22} & \textbf{0.627}          & \textbf{0.792} & \textbf{8.20} & \textbf{0.180}          & \textbf{0.423} \\ \midrule
YieldNet~\citeyearpar{khaki2021simultaneous}                & 30.96          & 0.015                   & 0.124          & 87.41         & 0.095                   & 0.310          & 9.23          & 0.003                   & 0.060          & 9.46         & 0                   & 0          \\
PhenoYieldNet-MC (Ours)            & \textbf{16.52} & \textbf{0.516}          & \textbf{0.718} & \textbf{54.88} & \textbf{0.638}          & \textbf{0.799} & \textbf{5.91} & \underline{0.616}          & \underline{0.785} & \underline{8.32} & 0.136          & 0.368 \\ \bottomrule
\end{tabular}
\caption{
Performance comparison of yield prediction on CropNet dataset. Cotton yield is in pounds per acre (lb/ac); other crops are in bushels per acre (bu/ac).
The best and second-best results are highlighted in \textbf{bold} and with an \underline{underline}, respectively.
}
\label{table_cropnet_baseline}
\end{table*}

\subsubsection{Evaluation Metrics}
We adopt three evaluation metrics for comprehensive assessment: Root Mean Square Error (RMSE) to measure the absolute prediction error, R-squared ($R^2$) to evaluate the goodness of fit, and the Pearson Correlation Coefficient (Corr) to quantify the linear relationship between predicted and actual values.
Formal definitions of these metrics are provided in Supplementary Material.

\subsubsection{Implementation Details}
The proposed PhenoYieldNet adopts SpectralGPT as the backbone encoder.
For pre-training, PhenoYieldNet is trained for 200 epochs using the AdamW optimizer~\cite{loshchilov2017fixing}, with $\beta_1 = 0.9$, $\beta_2 = 0.95$, and a weight decay of 0.05. A cosine learning rate decay schedule~\cite{loshchilov2016sgdr} is applied, starting from an initial learning rate of $1\mathrm{e}{-4}$, preceded by a warm-up phase of 20 epochs.
After pretraining, we fine-tune PhenoYieldNet for an additional 100 epochs with early stopping criteria, and it is optimized using AdamW with $\beta_1 = 0.9$, $\beta_2 = 0.999$, a cosine decay schedule, an initial learning rate of $3\mathrm{e}{-4}$, and 5 warm-up epochs.
The number of crop types $C$ is set as 4 for CropNet, while the CPB module is not employed for the single-crop dataset MODIS.
And the temporal masking for growing stages is applied at stage II to $\mathbf{Z}$, just before it is fed into decoder.
All experiments are implemented in PyTorch and conducted on an NVIDIA A6000 GPU.

\subsection{Comparison with Baselines}

\subsubsection{Single-Crop Yield Prediction}
We first evaluate PhenoYieldNet and other single-crop yield prediction methods, including MMST-ViT, Transformer, UNet-ConvLSTM, and MMVF, on the MODIS dataset.
As shown in Table~\ref{table_modis_baseline}, PhenoYieldNet consistently outperforms all competing methods.
For example, compared to the second-best method (i.e., UNet-ConvLSTM), our approach reduces RMSE by 0.38 and improves $R^2$ by 0.077.

We further evaluate PhenoYieldNet on the CropNet dataset by comparing it with other baseline methods. Specifically, PhenoYieldNet is trained individually for each crop (denoted as PhenoYieldNet-SC), and the results are reported in the first section of Table~\ref{table_cropnet_baseline}.
The results show that PhenoYieldNet consistently achieves the most accurate yield predictions, with the lowest errors and the highest correlation scores.
Specifically, for \textit{soybean}, our method achieves the lowest RMSE of 6.22 and the highest correlation metrics, with $R^2$ = 0.627 and Corr = 0.792.
For \textit{corn}, compared to the second-best method (i.e., RF), PhenoYieldNet reduces RMSE by 4.69 and improves $R^2$ and Corr by 0.085 and 0.038, respectively.
These superior results demonstrate that even under a single-crop setting, the combination of our pretraining strategy and our CPA module significantly enhances prediction accuracy across different crops and regions.

\begin{figure}[t]
\centering
\includegraphics[width=1.0\linewidth]{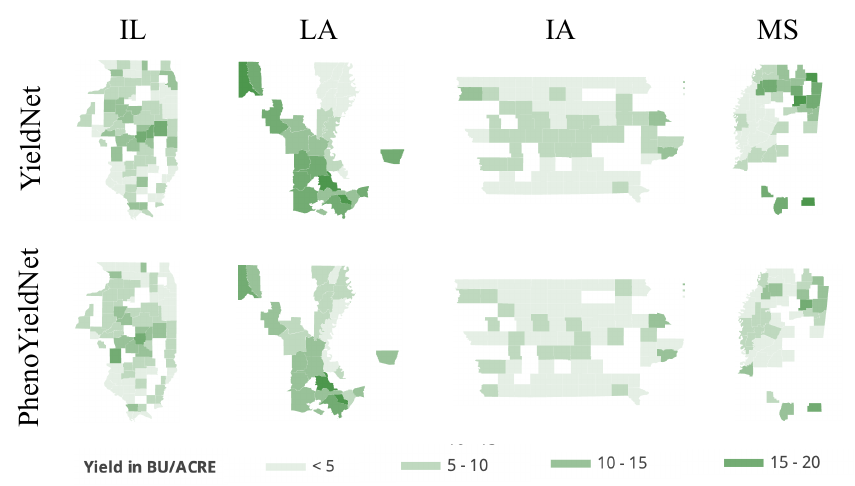}
\caption{Yield prediction error distribution for \textit{corn} across four U.S. states.}
\label{Fig3}
\end{figure}




\subsubsection{Multi-Crop Yield Prediction}
In the multi-crop prediction scenario, PhenoYieldNet (denoted as PhenoYieldNet-MC, with $C=4$) is compared with YieldNet. As shown in Table~\ref{table_cropnet_baseline}, PhenoYieldNet consistently achieves lower RMSE and higher $R^2$ and Corr scores across all four evaluated crops, indicating superior robustness and adaptability to diverse phenological patterns.
YieldNet, in contrast, performs poorly—particularly in $R^2$ and Corr—potentially due to its CNN-based architecture, which struggles to capture temporal variations in crop growth and tends to produce low-variance predictions, resulting in diminished correlation with ground-truth values. It is worth noting that a performance trade-off is observed for \textit{winter wheat} and \textit{soybean} in terms of $R^2$ and $Corr$ when comparing to PhenoYieldNet-SC. 
This may be due to class imbalance, as winter wheat has significantly fewer samples. Another contributing factor is its unique phenology: winter wheat exhibits a more atypical phenology with a full-year growing cycle, whereas other crops are predominantly summer-planted, making its phenological dynamics harder to model.  Consistent with this explanation, other methods also show degraded performance on winter wheat.

To further demonstrate model performance, we visualize the prediction errors of PhenoYieldNet and YieldNet across various crops in Fig.~\ref{Fig3}. Lighter colors indicate lower errors, highlighting PhenoYieldNet’s effectiveness in reducing extreme prediction deviations. Additional visualization results are provided in Supplementary Material.

\subsection{Ablation Studies}

We conduct a comprehensive ablation study to analyze the effects of PhenoYieldNet's key components, including pre-trained foundation model knowledge, proposed TCA strategy and crop-aware temporal decoder (CPB and CPA modules), on the CropNet dataset. All the results are in Table~\ref{tab:ablation_unified}.

\subsubsection{Effect of RSFM Knowledge and TCA Strategy.}
We first analyze the effect of foundation model knowledge by comparing the model (1) training from scratch, and model (2), in which the encoder is initialized with a general-purpose RSFM. It can be observed that directly fine-tuning model (2) results in performance degradation on crops such as \textit{cotton} and \textit{soybean}, despite achieving significant gains on \textit{corn} and \textit{winter wheat}. This is aligned with our assumption that there exists a domain gap when directly adapting pretrained RSFM without incorporating agricultural knowledge. And it also motivates the introducing of TCA strategy to mitigate the domain gap while addressing generalization ability.

Next, we assess the impact of proposed TCA strategy by comparing the model (2) to the full PhenoYieldNet. As shown, applying TCA yields consistent improvements across all the crop types. 
Although compared to the model (1), there is a subtle performance trade-off on the dominant crop (i.e., \textit{soybean}). We attribute it to the sensitivity of fundamental knowledge differences, while model (1) potentially overfits on agriculture knowledge. Yet, the PhenoYieldNet still achieves the best overall performance across all crop types.


\subsubsection{Effect of CPB Module.}

We compare model (3), a single-crop baseline trained independently for each crop type, with model (4), which incorporates a Crop Phenology Bank (CPB) into the same architecture to support unified yield prediction across multiple crops.
The results demonstrate that the CPB module brings consistent improvements across most of the crop types, which suggests that enabling the model to differentiate between crops via a set of learnable queries is effective for multi-crop yield prediction. Although there is a slight performance degradation for \textit{soybean} in terms of RMSE, which is also observed when comparing full PhenoYieldNet with its single-crop counterpart, PhenoYieldNet-SC. 
We attribute this to the joint feature learning with a shared encoder, which may slightly compromise specialization for a few crops while benefiting the generalization to others.


\begin{table*}[tp]
\centering
\setlength{\tabcolsep}{0.85mm}
\begin{tabular}{lcccccccccccc}
\toprule
\multicolumn{1}{l}{\multirow{2}{*}{Model Configuration}} & \multicolumn{3}{c}{Corn} & \multicolumn{3}{c}{Cotton} & \multicolumn{3}{c}{Soybean} & \multicolumn{3}{c}{Winter Wheat} \\
\cmidrule(lr){2-4} \cmidrule(lr){5-7} \cmidrule(lr){8-10} \cmidrule(lr){11-13}
& RMSE $\downarrow$ & $R^2$ $\uparrow$ & Corr $\uparrow$ & RMSE $\downarrow$ & $R^2$ $\uparrow$ & Corr $\uparrow$ & RMSE $\downarrow$ & $R^2$ $\uparrow$ & Corr $\uparrow$ & RMSE $\downarrow$ & $R^2$ $\uparrow$ & Corr $\uparrow$ \\
\midrule
(1) Train from scratch & 21.28 & {0.363} & 0.603 & 67.32 & 0.571 & 0.755 & \textbf{5.76} & \textbf{0.621} & \textbf{0.788} & 9.70 & 0.035 & 0.185   \\
(2) w/ RSFM pretraining  & 18.23 & 0.331 & 0.575  & 86.22 & 0.062  & 0.248& 6.61 & 0.550  & 0.741 & \underline{8.38} & 0.080& 0.283 \\
(*) PhenoYieldNet &  \textbf{16.52} & \textbf{0.516} & \textbf{0.718} & \textbf{54.88} & \textbf{0.638}  & \textbf{0.799} & \underline{5.91} & \underline{0.616} & \underline{0.785} & \textbf{8.32} & \underline{0.136} & \underline{0.368} \\ 
(3) w/o CPB \& CPA & 17.93  &  0.365  &  0.604 &  71.39  & 0.418   & 0.647   &  6.25  &  0.507  &  0.712 & 8.73  & 0.102 & 0.321 \\
(4) w/o CPA & \underline{17.83} & \underline{0.482} & \underline{0.694} & \underline{61.60} & \underline{0.577} & \underline{0.759} & 6.82 & 0.551 & 0.742 & 8.47 & \textbf{0.151} & \textbf{0.388} \\ 
\bottomrule
\end{tabular}
\caption{Ablation study on PhenoYieldNet's key components. 
The best and second-best results are highlighted in \textbf{bold} and with an \underline{underline}, respectively.
}
\label{tab:ablation_unified}
\end{table*}

\subsubsection{Effect of CPA Module.}
We evaluate the contribution of the CPA module by comparing model (4) with the full PhenoYieldNet. Model (4) is constructed by replacing the CPA module with a standard temporal attention mechanism.
As shown, incorporating CPA leads to consistent performance improvements across all crop types.
For example, the RMSE for \textit{cotton} decreases from 61.60 to 54.88, and the $R^2$ increases to 0.638.
These results suggest that explicitly modeling temporal trends and variations through CPA enhances yield prediction accuracy.
To further understand the mechanism of the CPA module, we visualize the attention maps and crop growing curves to examine the contributions of different crop growth stages. Details are provided in Supplementary Material.

\begin{figure}[t]
\centering
\includegraphics[width=1.0\linewidth]{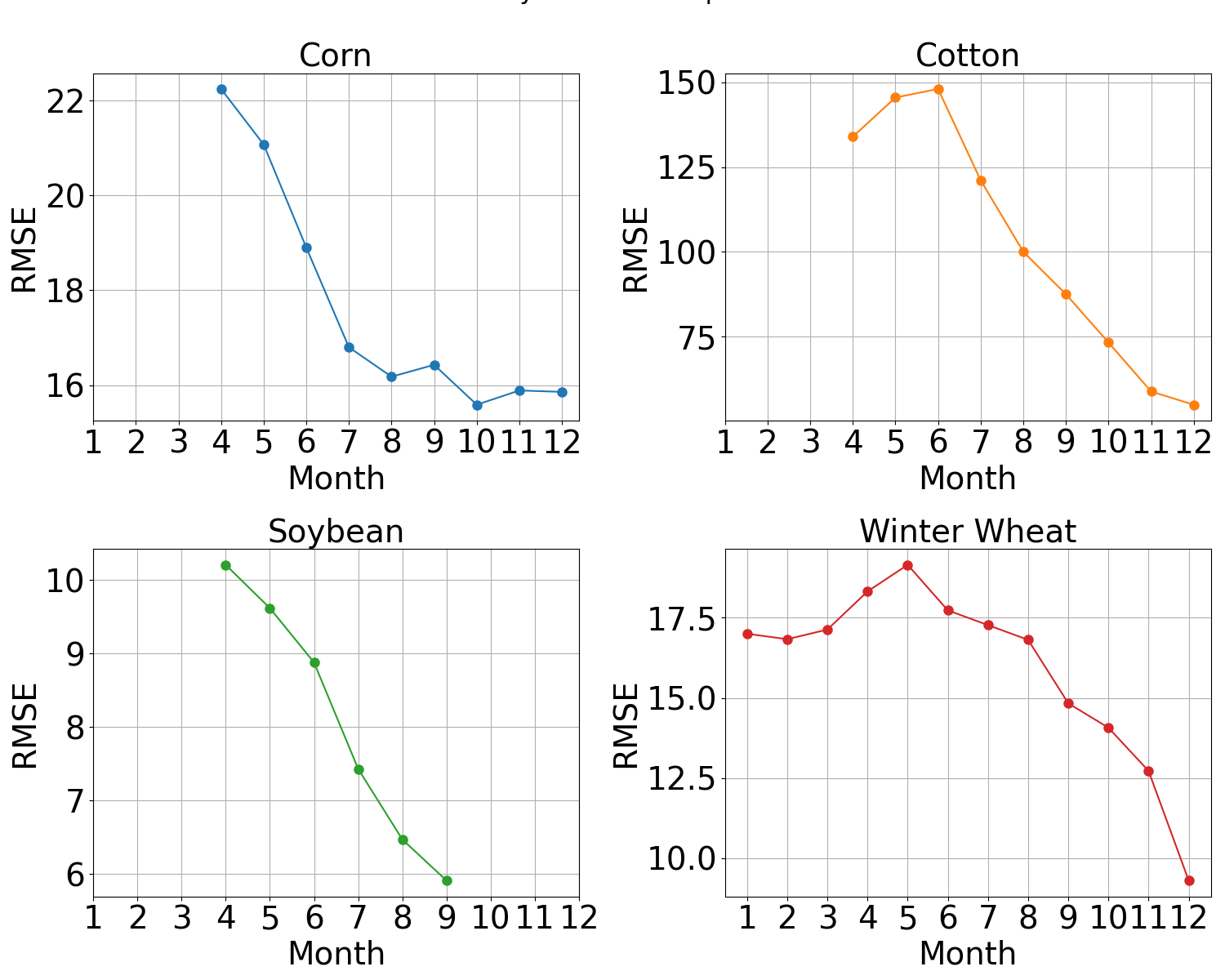}
\caption{RMSE of real-time multi-crop yield prediction.}
\label{real-time}
\end{figure}

\begin{figure}[t]
\centering
\includegraphics[width=1.0\linewidth]{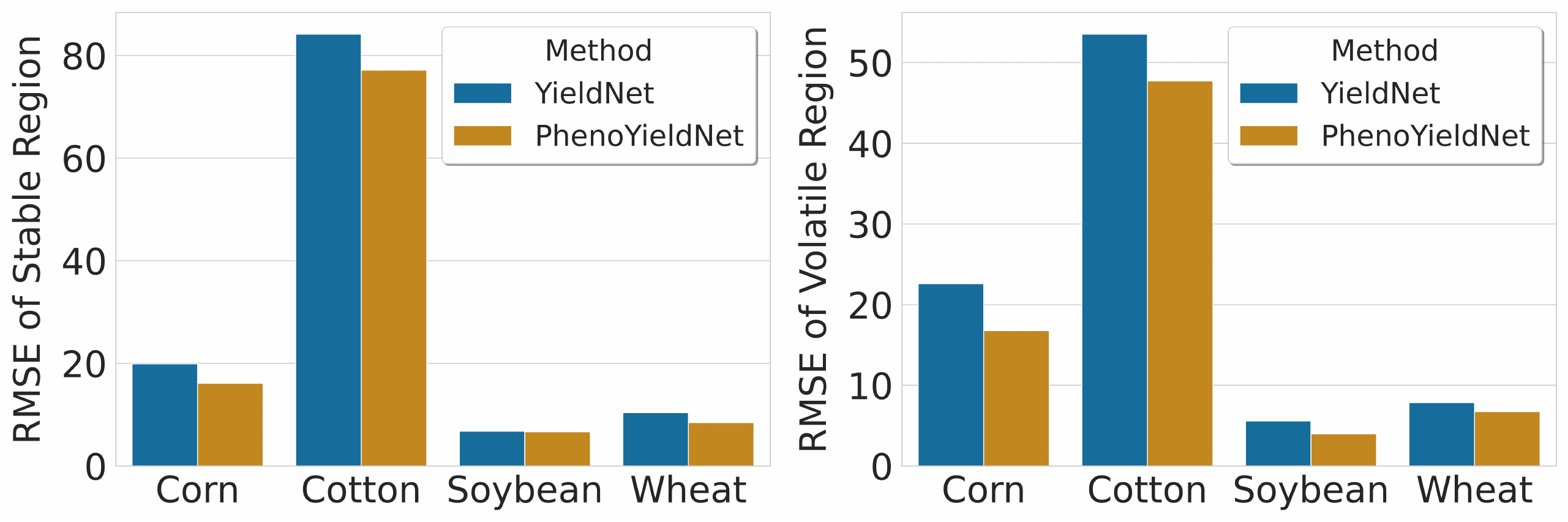}
\caption{RMSE in stable vs. volatile weather regions.}
\label{robust}
\end{figure}

\subsection{Discussions}

\paragraph{Real-Time Yield Prediction.}

Given that different crops exhibit distinct phenological characteristics, we conduct an real-time prediction experiment by progressively predicting crop yields based on the input sequences evolving across the growing season, i.e., $\mathbf{X}^{SITS}_{sos:t}$ and $\mathbf{X}^{MTS}_{sos:t}$ for $t \in \{sos, \dots,eos\}$, with the data for each crop commencing from start-of-season~($sos$) to end-of-season~($eos$).
As shown in Fig.~\ref{real-time}, prediction accuracy generally improves as more observations become available, reflecting the model’s ability to incorporate informative phenological signals. 
Interestingly, we observe subtle fluctuations in RMSE for certain crops. In the case of \textit{winter wheat}—which is typically planted earlier than others—the RMSE decreases early in the season as expected, but a temporary increase is observed around April. This may be due to changes in the shared representation space, as the model begins integrating signals from newly emerging crops planted later in the season.


\subsubsection{Robustness to Weather Variation}

To assess PhenoYieldNet under varying weather conditions, we divide the test set into stable (bottom 70\% in the coefficient of variability of meteorological variables) and volatile regions (top 30\%).
As shown in Fig.~\ref{robust}, all methods exhibit lower RMSE in stable regions compared to volatile ones.
PhenoYieldNet shows marginal improvements in stable regions but achieves significantly better performance in volatile regions, indicating its superior robustness to weather variability.

\subsection{Limitation}
While PhenoYieldNet unifies the prediction of multiple crops and offers significant efficiency and scalability, two limitations remain.
First, the crop phenology bank is constructed based on crop species. A more flexible design at the unit or elemental level—such as using vector quantization to adaptively select codes across different crops—could enhance the model’s ability to generalize to unseen species beyond the training domain. 
Secondly, the multi-crop training paradigm is sensitive to distributional shifts and class imbalance, resulting in a slight performance gap for minority crops.
These trade-offs highlight opportunities for improving the model’s generalization, efficiency, and robustness in future research.
\section{Conclusion}

We present PhenoYieldNet, a unified framework for multi-crop yield prediction that models crop-aware phenological responses to long- and short-term meteorological variations.
A crop-aware decoder integrates a crop phenology bank to encode crop-specific knowledge and a crop phenology attention module to capture interactions with meteorological signals.
PhenoYieldNet is trained in two stages: temporal contrastive adaptation for domain alignment with remote sensing knowledge, followed by decoder fine-tuning.
Experiments show that PhenoYieldNet outperforms state-of-the-art methods in both single- and multi-crop settings.


\section*{Acknowledgements}
This work was supported by the SmartSat CRC, whose activities are funded by the Australian Government’s CRC Program, and by the ECU Early-Mid Career Researcher (EMCR) Grant Scheme.

{
    \small
    \bibliographystyle{ieeenat_fullname}
    \bibliography{main}
}


\end{document}